\documentclass{article}

\usepackage{arxiv}

\usepackage[utf8]{inputenc} 
\usepackage[T1]{fontenc}    
\usepackage{hyperref}       
\usepackage{url}            
\usepackage{booktabs}       
\usepackage{amsfonts}       
\usepackage{nicefrac}       
\usepackage{microtype}      
\usepackage{lipsum}
\usepackage{graphicx}

\title{Opportunities and challenges in partitioning the graph measure space of real-world networks}

\author{
 Máté Józsa \\
  Department of Physics\\
  Babeş-Bolyai University\\
  M. Kogălniceanu nr. 1, 400084, Cluj-Napoca, Romania \\
   \And
 Alpár S. Lázár \\
  Faculty of Medicine and Health Sciences\\
  University of East Anglia\\
  NR4 7TJ, Norwich, UK \\
  \And
 Zsolt. I. Lázár\\
  Department of Physics\\
  Babeş-Bolyai University\\
  M. Kogălniceanu nr. 1, 400084, Cluj-Napoca, Romania \\
  \texttt{zsolt.lazar@ubbcluj.ro} \\
}

\begin{document}
\maketitle
\begin{abstract}
Based on a large dataset containing thousands of real-world networks ranging from genetic, protein interaction, and metabolic networks to brain, language, ecology, and social networks we search for defining structural measures of the different complex network domains (CND).
We calculate 208 measures for all networks and using a comprehensive and scrupulous workflow of statistical and machine learning methods we investigated the limitations and possibilities of identifying the key graph measures of CNDs.
Our approach managed to identify well distinguishable groups of network domains and confer their relevant features. These features turn out to be CND specific and not unique even at the level of individual CNDs.
The presented methodology may be applied to other similar scenarios involving highly unbalanced and skewed datasets. 
\end{abstract}


\section{Introduction}
It is becoming a popular belief that the present century is about complexity \cite{bib1}.
Network science appears to be the philosopher’s stone of complex systems as they not only offer important insights into previously intractable systems but they also represent the primary tool in answering highly critical questions ranging from human resource management in a company, to cascading power network failure or drug design \cite{bib2, bib3}.
In its basic form the network model is made of elements of utter simplicity, nodes representing constituents or states of the system and edges depicting interactions, relationships or processes involving two such elements.
These simple ingredients can give rise to practically limitless configurations,  referred to as the network’s topology, which can be mapped to concrete or abstract realizations of systems and phenomena. 
In recent years the number of network measurables has grown steadily, propelled by the fast paced buildup in the corpus of available data \cite{bib4} backed up by the unyielding Moore’s law and equivalents on computational, transmission and storage capabilities.
Node degree, the  number of first neighbors reachable from a given node, and the large scale properties of its distribution are sufficient for grasping the essence of certain types of network and the systems they model.
Universal laws governing the evolution of widely different systems have been discovered relying on this simple measure \cite{bib5, bib6, bib7, bib8}.
Exploring the role of cliques, i.e., fully connected subgraphs, appears to be the right approach in studying for example the structure and functioning of the human connectome \cite{bib9, bib10}. The quantitative description of other systems requires network measures of varying degree of complexity and intuitiveness.
Prominent examples include concepts like small-world-ness of social and web networks \cite{bib4, bib11}, modularity, hierarchical clustering, special motifs and robustness in the study of cellular networks, like metabolism or protein-protein interactions \cite{bib12, bib13, bib14, bib15, bib16, bib17, bib18}, as well as cycles and k-core distributions responsible for the stability of the vascular system of leaves, different ecological systems or financial networks \cite{bib19, bib20, bib21, bib22}.
Some dynamical processes on networks can be mapped to synchronization phenomena \cite{bib23, bib24}, which can be characterized by the ratio of extremal values of the Laplacian matrix of the network.
The interplay between global, local and asynchronous phases are influenced by the degree correlations in the network which also affects its controllability and percolation properties \cite{bib25, bib26, bib27, bib28, bib29, bib30, bib31, bib32}.
Epidemic, rumor and information spreading in biological, social and technological networks are affected predominantly by the properties of the underlying connectome \cite{bib33}.
For instance, the leading eigenvalue of the probability transition matrix is related to the speed of the dynamics on the network \cite{bib34}.
Different centrality measures were proposed to identify important nodes of the networks in various scenarios \cite{bib35, bib36}.
An important body of research involving network science targets the understanding of mechanisms behind the formation, and evolution of real networks.
A typical approach starts off from common sense premises on the relevant elements and processes of the modelled complex systems and assumes some foreseeable effects on various network measures.
In many cases, however, it is not trivial, which are the network features to look at.
It is also common practice to compare real networks to null models created by constrained randomization of the network \cite{bib37}.
Nonetheless, due to the inherent biases of the different techniques the results appear to tell us not so much about the networks themselves as about the interaction of the randomization process with the particular network topology that is being investigated.
Recent advances in creating null models, such as in \cite{bib38, bib39}, avoid blatant biases given a specific constraint, like keeping the degree sequence, e.g., with the configuration model, the Havel-Hakimi algorithm or degree preserving randomization \cite{bib35, bib40}.
Yet the question remains  as to which is the right  set of measures to be kept constant in the different real-world scenarios.
In summary all the above mentioned methods try to capture specifics of complex network domains (CNDs) without comparing them directly with other CNDs.
Our work is based on the view that in order to exploit the potential of network theory in the study of a complex system one first has to identify the distinctive properties of the underlying network.
In other words connections between the system’s defining properties and its network measures should be formulated in light of the relationships between other classes of real networks and the same network measures.
Here we propose to identify the network measures that best define the different types of real networks.
These expectations were  also formulated by Costa et al \cite{bib41}.
There the authors offered detailed instructions on the classification procedure in terms of the  data analysis methods popular at the time.
Later research shows that there are network measures that can identify certain types of real-world networks with a relatively high accuracy \cite{bib42, bib43}.
Rossi and Ahmed \cite{bib42} investigated 530 real-world networks along with 483 synthetic networks, using 11 graph measures.
They concluded that in general networks are already distinguishable using only three or four measures.
Another work is based on 986 real-world networks and 575 generated ones and 8 descriptive features \cite{bib43}.
There, the authors determined pairs of measures that optimally separate the different types of network domains based on the feature importances returned by the Random Forest supervised classification algorithm. 
In this work we present a comprehensive analysis which aims to provide a tool for answering the question to what extent and under what circumstances there is an optimal parameter space of the real-network that one wants to examine.
Our focus is the space of those network structure measures wherein CNDs form distinguishable regions.
Following Rossi and Ahmed \cite{bib42} our primary indicator will be the performance scores of supervised classification models.
We emphasize, however, that “recognizing” network domains represents the means rather than the goal.
Therefore the study is limited to simple graphs and non-trivial network measures.
By trivial measure we mean one that is obvious to the naked eye, e.g., most of the brain networks in the data set are excessively large while the edge density of road networks are conspicuously below average due to their geometric nature.
For the same reason our networks are stripped down to undirected and unweighted edges as these extra degrees of freedom would shadow the structural properties of interest.
Generated (random) networks such as Erdős-Rényi or Barabási-Albert are shown to be easily distinguishable from real networks \cite{bib42}.
Thus present work will not cover them. 
Instead of arbitrarily picking fashionable network measures we employ a systematic methodology for discovering these.
Also we use an up-to-date machine learning methodology in order to avoid several pitfalls.
We attempt to lay out a route that can help those struggling with such highly unbalanced and skewed data as that of graph measures in a real-world network scenario.
In the following the terms CND and (network) domain, group or class will be used with equivalent denotation.
Similarly graph measures will be also referred to as network features or structural properties.

\section{Methodology}
\label{sec:headings}
In order to extract the relevant properties of CNDs we pursued the following scheme: collect raw network data from public repositories, standardise them by applying a systematic preprocessing procedure then select the relevant features specific to each CND by the machine learning framework developed for this purpose based on  consecutive filtering  and wrapping procedures (See Section \ref{S:filtering} and \ref{S:wrapping}) \cite{bib44}.

\subsection{Raw network data}
\begin{figure}	
    \centering
	\includegraphics[width=\textwidth]{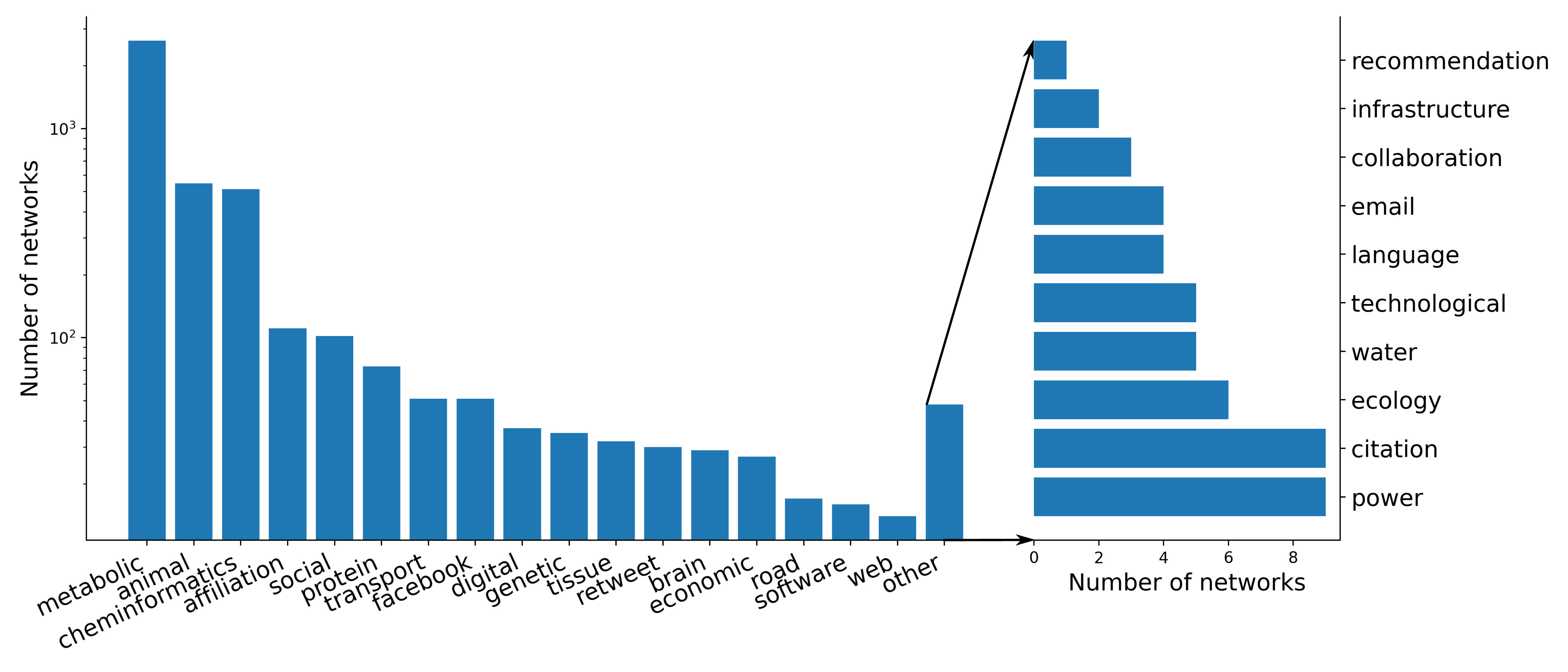}
	\caption{Number of networks used in the study. Groups with less than 10 elements are collected under the name “other”.
	Its content is detailed on the right hand side of the figure.
	The meaning and origin of different network groups are explained in Table S1 in the Supplementary Material.}
	\label{F:fig1}
\end{figure}

We compute 208 different topological network measures on 4367 real-world networks from 27 different domains (Figure \ref{F:fig1}).
These networks originate from Network Repository (\url{http://networkrepository.com/}) \cite{bib45}, BioGRID (\url{https://thebiogrid.org/}) \cite{bib46}, BioModels (\url{https://www.ebi.ac.uk/biomodels/}) \cite{bib47}, transport networks (\url{http://transportnetworks.cs.aalto.fi/}) \cite{bib48} and the CommunityFitNet corpus \cite{bib49} from the Index of Complex Networks (ICON) (\url{https://icon.colorado.edu/}).
Groups with less than 10 elements were aggregated into the group named “other”.
The collected and preprocessed data is made available online at \url{https://github.com/MateJozsaPhys/CNDinvestigation}.
Data sources, algorithms and missing values are indicated therein.
The highly unbalanced nature of the dataset can be observed by comparing the broad range of sizes on the logarithmically scaled axis in Figure \ref{F:fig1}.

\subsection{Preprocessing}
We converted the collected networks to undirected, unweighted graphs, without parallel edges or self-loops and only retained the giant component.
Bipartite graphs were replaced by their one-mode projections.
These simplifications offered multiple benefits.
In line with our main objectives it preserved only information on their “pure” non-trivial topology while extending the range of applicable network measures.

\subsection{Quantitative analysis}
\begin{figure}	
    \centering
	\includegraphics[width=\textwidth]{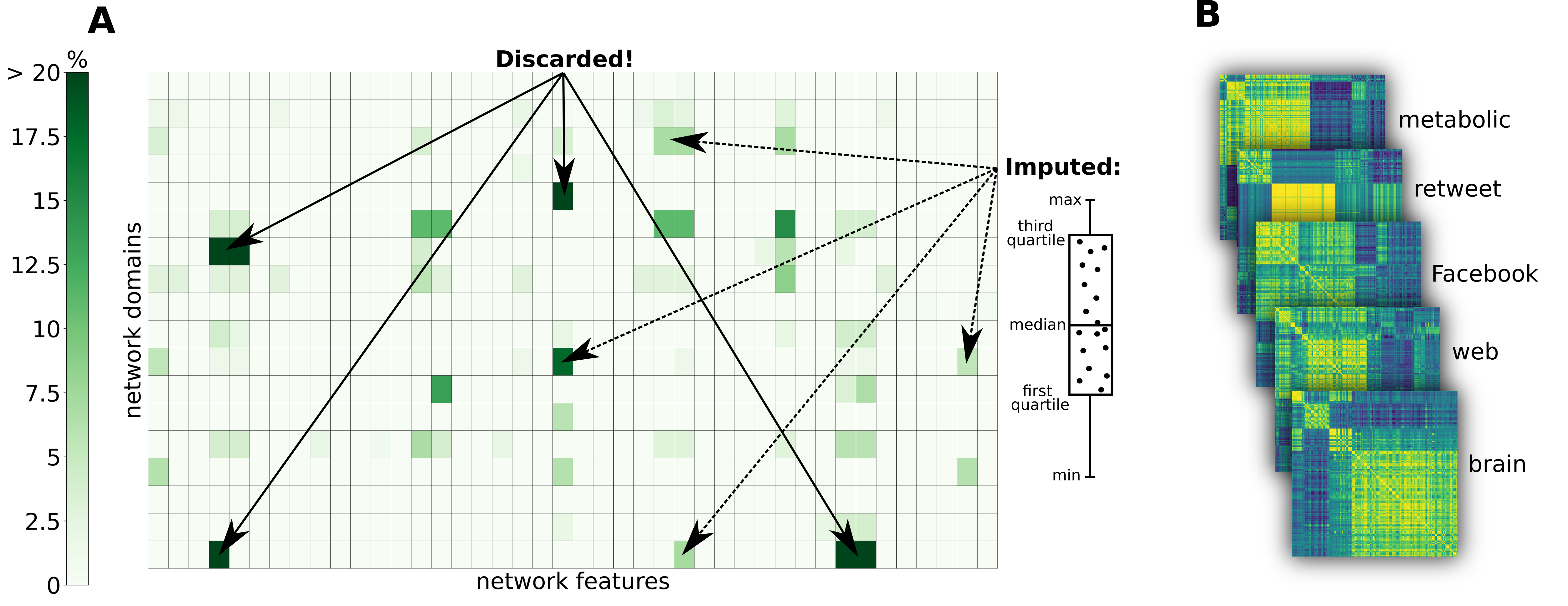}
	\caption{A) Illustration of handling missing data in the dataset.
	The lightness of the color of squares associated with a domain-feature pair represents the ratio of missing values.
	Features not calculated in more than 20$\%$ in the given group were discarded.
	In all other cases missing values were replaced by uniformly randomized values between the first and third quartiles of the distribution of valid values for the given group-feature pair.
    B) Feature-to-feature Pearson correlation matrices showing the differences in correlations among all 208 features. Lighter colors represent higher correlation coefficients.
    Each correlation matrix corresponds to a CND and is clustered by hierarchical agglomeration. One can notice the fundamental differences in the correlation patterns by comparing the sizes and distribution of clusters.}
	\label{F:fig2}
\end{figure}

We limited ourselves to network measures that are easier to implement and computationally not overly costly.
The 208 calculated topological measures include both global and local properties.
The node/edge-level measures were transformed to global measures by taking their average, minimum-, maximum values and their moments up to order four.
Moments were calculated after normalizing the distribution by dividing it by its maximum value. The computational complexity of measures depends on a variety of topological descriptors, like the number of nodes, density of edges, or the average of path lengths, and so on \cite{bib36}.
By upper limiting  memory usage and computational time a valid measure was returned in 99,8$\%$ of the runs. 
Networks missing more than 20$\%$ of the 208 measures were removed from the dataset.
Similarly, features not calculated for at least 80$\%$ of the networks in a domain were discarded altogether for the given domain (Figure \ref{F:fig2}A).
In the other cases missing values were compensated for by imputation on a domain-by-domain basis. It is common to use a fixed value such as the domain mean or the median \cite{bib50, bib51}. However, in our case this method would produce spurious markers of the domain that would mislead the classification algorithm and would make the identification of the group trivial.
Therefore missing values were replaced by random numbers distributed uniformly between the first and third quartiles of the valid values for the particular domain.
Given the very small ratio of missing values their effect is negligible. 
Also, to avoid the above mentioned trivial classifications along constant dimensions, we removed all features that had more than 80$\%$ constant values inside a CND.
As a result the number of features was reduced to 164. Features which were constant in more than 80$\%$ of cases, for instance structural holes, were also discarded \cite{bib4}.
The measures calculated  in the study are summed up in Table S2 in the Supplementary Material, explaining the abbreviations and providing information on their implementations.

\subsection{Feature selection}
The number of network measures and that of the networks is comparable requiring a drastic reduction in the former.
Because of the intrinsic connections between  measures, for instance, cliques occur in larger size and number in denser networks, and for better interpretability the number of features again has to be reduced.
To that end we applied two subsequent  fundamental feature selection methods.
First the computationally  less expensive filtering followed by wrapping.
While filter methods characterize the intrinsic properties of data in the absence of a classifier, wrapper methods measure the relevance of features based on the performance of a supervised classifier \cite{bib44}.
By choosing appropriate learning algorithms the negative effect of correlations in the data can, to a large extent, be mitigated (see Section \ref{S:wrapping}).
However, reducing correlations is also an important means in finding dominant topological features.
Here we consider both linear and non-linear correlations as intrinsic indicators for the filtering (see Section \ref{S:filtering}) and the Random Forest classification model will constitute the core of the wrapper.

\subsubsection{Filtering}
\label{S:filtering}
Interestingly, despite the general mathematical connections between network features the correlations  exhibit no universality.
The feature-to-feature Pearson correlation patterns in Figure \ref{F:fig2}B show fundamental differences across groups.
Therefore the filtering was made separately for each group of networks.
Linear correlations were reduced by removing one from each pair of features with a coefficient bigger than 0.9.
Subsequently, non-linear correlations were cut down by repeating the above procedure using  Spearman correlation coefficients.
The number of retained features was between 21 and 48 depending on the group.

\subsubsection{Wrapping}
\label{S:wrapping}
\begin{figure}	
    \centering
	\includegraphics[width=\textwidth]{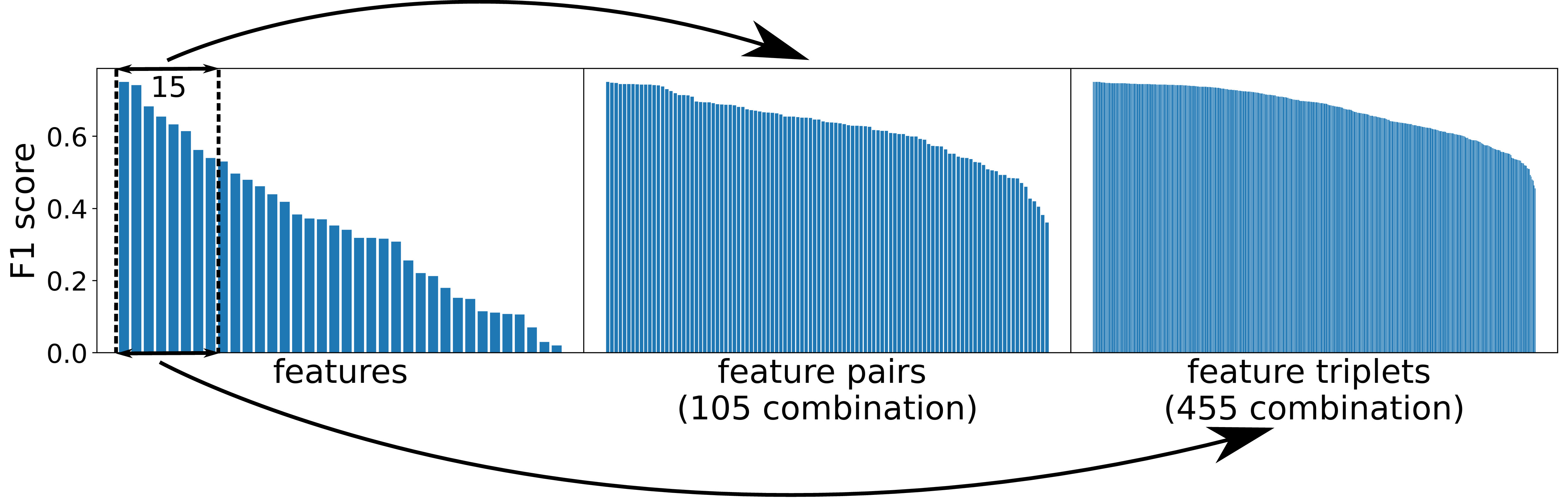}
	\caption{Illustration of the modified forward selection procedure. First, the model is fitted with each of the features one by one  (left panel).
	The best performing 15 features are combined into  all possible doublets and triplets.
	The model is then  fitted  separately with each of the proposed combinations of features.  Cross validation is repeated three times and averaging over all splits is applied.
	The figure illustrates the case of affiliation networks.}
	\label{F:fig3}
\end{figure}

Consistent with our objective to discover the distinguishing properties of a CND from others we applied a One-vs-Rest classification scheme.
After testing  a variety of competing supervised learning classifiers including Logistic Regression, Support Vector Machine and K-Nearest Neighbour classifier we conclude that Random Forest is the most suitable for a highly nonlinear and unbalanced One-vs-Rest classification setup applied here \cite{bib52, bib53, bib54, bib55}.
The parameters of Random Forest could be relatively easily tuned for optimal behaviour as opposed to other models.
In order to avoid overfitting  the maximum depth of a tree was set to three.he number of trees in the forest were chosen to be 100.
Balancing over subsamples of trees was applied to moderate the effect of class imbalances (see Figure \ref{F:fig1}).
As a check we repeated the classification procedure using undersampling, i.e., including only a subset of elements (networks) from the larger classes (CNDs). In most cases we obtained only small perturbations in the results ruling as unjustified the loss of valuable data due to undersampling. The performance of the classifier was estimated through the F1 score, the harmonic mean of precision and recall values, being a good alternative for the popular ROC AUC (Area Under the Receiver Operating Characteristic Curve) score which appears not to capture the true performance of the minority class (the “One” in the “One-vs-Rest”) in highly unbalanced scenarios \cite{bib56}.
To improve the generalization power of our model the F1 scores were calculated applying repeated stratified k-fold cross validation, with five folds, repeated three times \cite{bib54}.
Let us recall the original motivation behind the pursued classification task, namely, obtaining a handful network measures that may help in understanding the modelled complex system.
As such we limit our query to no more  than three distinguishing features per CNDs. To take advantage of the strength of both popular feature extraction methods, i.e. the bottom-up forward selection and the top-down backward elimination, we used a new, modified forward selection \cite{bib44}.
The method used here differs from the original forward selection method in that it does not derive feature triplets from pairs, rather both are constructed from the first 15 elements based on their F1 scores in one dimension (Figure \ref{F:fig3}).
As a consistency check and a study of the modified forward selection method we estimated  the overlap between the high performing feature pairs and triplets.
This was done as follows: the first 10 feature doublets were selected and completed to triplets, each pair being combined with the remaining 13 features, producing 130 triplets.
The size of the intersection of the so-obtained set with the best performing 130 triplets expressed in percentage can be interpreted as a measure for the consistency of the procedure (Figure \ref{F:fig5}A).

\subsection{The perspective of unsupervised learning}
\begin{figure}	
    \centering
	\includegraphics[width=\textwidth]{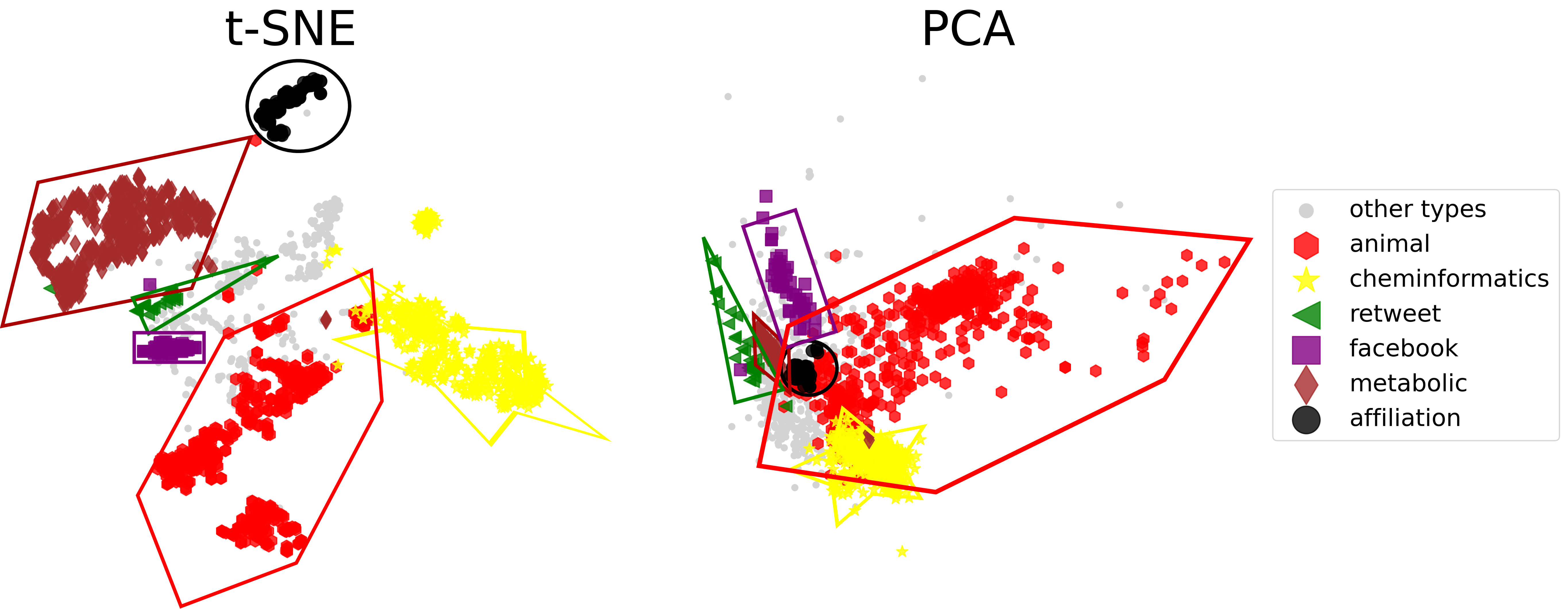}
	\caption{Illustration of the embedded and projected space from t-SNE unsupervised classification  and PCA into two-dimensions.
	Here the perplexity value used by t-SNE was 30. Also the elements of the groups were limited to 500 by random undersampling so that the dataset is not dominated by the largest group.
	The groups which are seemingly well separated  are marked with polygons.
	One can notice here that t-SNE can capture better the intrinsic properties of the data in contrast to PCA.
	This can be explained by the fact that the abundance of nonlinearities in the dataset makes the classification impossible with a linear model like PCA.}
	\label{F:fig4}
\end{figure}

One can discover intrinsic properties of the data also by applying dimensionality reduction techniques.
For a dataset that has so many nonlinear properties the unsupervised learning method t-Distributed Stochastic Neighbor Embedding (t-SNE) seems a more appropriate tool as opposed to the popular Principal Component Analysis (PCA) \cite{bib57, bib58}.
Figure \ref{F:fig4} demonstrates  the clear superiority of the nonlinear method.
On the downside, however,  one cannot interpret quantitatively the distances between groups in the embedded space.
Moreover the output  depends sensitively on the parameter known as  perplexity related to the average size of the clusters.
For a proper operation perplexity has to be tuned on a case by case basis. Given the above particularities of the unsupervised classification methods they did not contribute to the final results. 

\section{Results}
\begin{figure}	
    \centering
	\includegraphics[width=\textwidth]{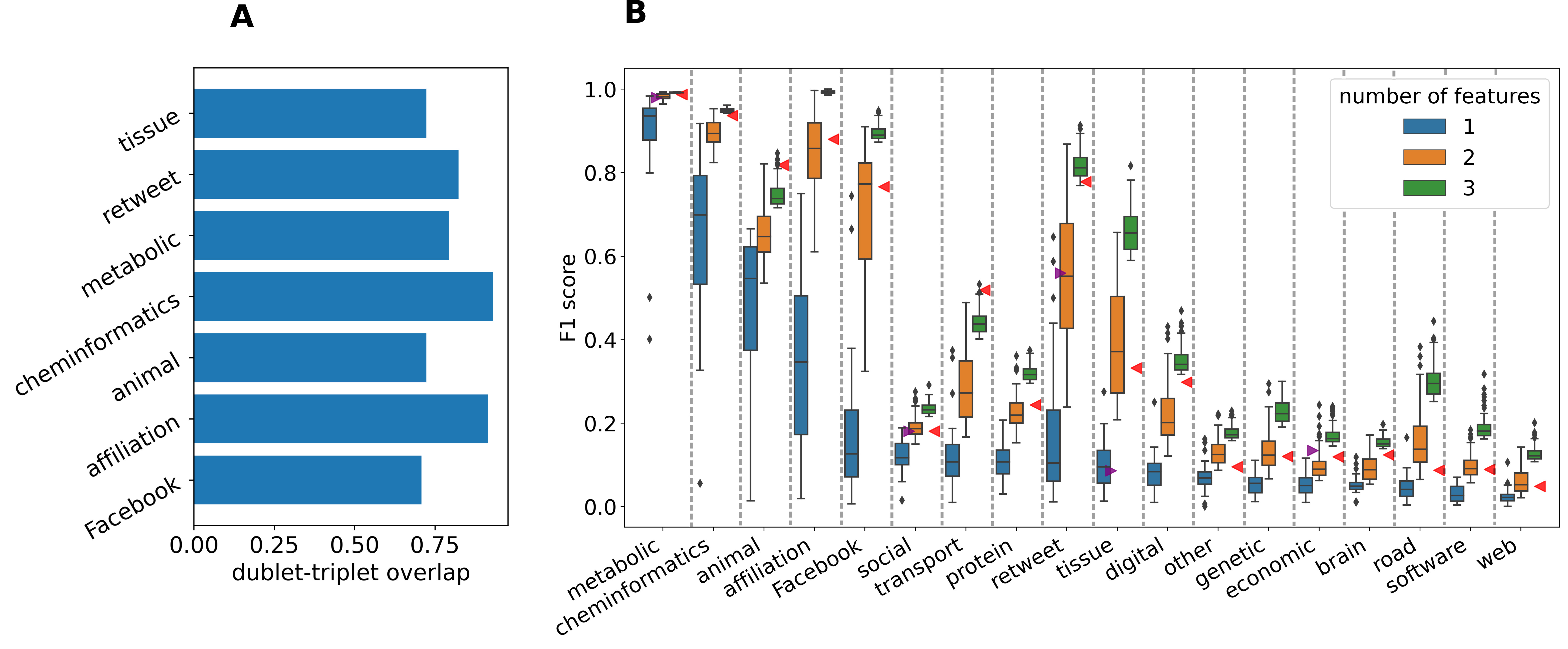}
	\caption{Results of the network classification in up to three dimensions according to the modified forward selection method described in Section \ref{S:wrapping}.
    A) Extent to which the best performing feature triplets encompass the best performing feature pairs.
    One can notice that the percentage of overlapping is dependent on the CNDs, indicating that the original forward feature selection would fail in some cases to determine the best feature triplets.
    B) Distribution of F1 scores resulting from the fitting with different combinations of one, two and three features. In the case of two and three-dimensions only the best 100 scores are represented.
    Red triangles represent the scores of the triplets used in \cite{bib42} while purple triangles mark the performance of feature pairs used in \cite{bib43}.
    The distributions are characterized by medians and quartiles, occasionally with outliers.
    The groups are arranged from left to right by the median of the F1 score of the  classification by a single feature.
    One can distinguish the “well-behaving” networks, which are separable in this scenario, i.e. the F1 score of classification exceeds 0.7 (well above random classification).}
	\label{F:fig5}
\end{figure}

\begin{figure}	
    \centering
	\includegraphics[width=\textwidth]{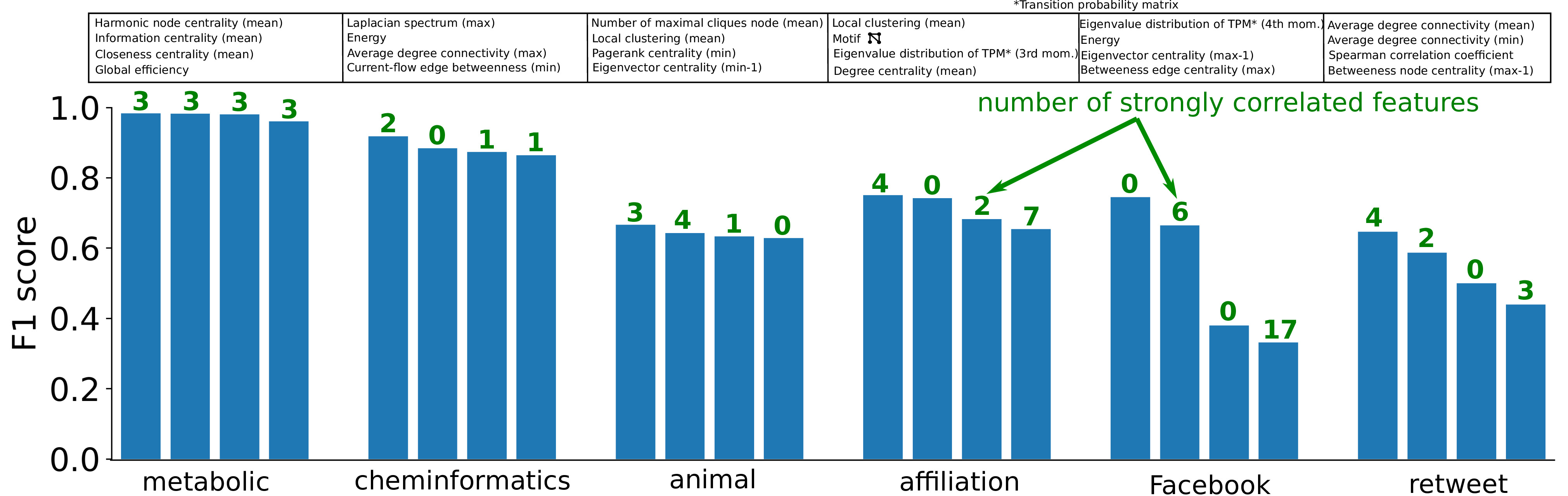}
	\caption{Distinguishing features of CNDs discernible by a single feature. Text boxes contain the names (top to bottom) of the defining feature singlets with F1 score bars below (left to right).
	Green numbers indicate the number of other features that highly correlate ($C_{Pearson} > 0.9$) with the nominated feature.
	Table S2 contains comprehensive details on all features including their implementation.}
	\label{F:fig6}
\end{figure}

Comparing the two-dimensional mapping of PCA and t-SNE on Figure \ref{F:fig4} suggests that the data has many nonlinear relations thus nonlinear  partitioning is to be favored over linear methods.
The results of the filtering described in Section \ref{S:filtering} show a strong domain dependence of the feature-feature correlations (see Figure \ref{F:fig2}B).
After reducing the number of strongly correlating ($C_{Pearson} > 0.9$) features we cannot come up with a feature set which even remotely suits all network groups.
Modest correlation values between the measures proposed in \cite{bib42, bib43} and the features emerging after our procedure suggest that indeed the set of relevant features can be  fundamentally different at the level of individual CNDs.
According to the nonlinear t-SNE unsupervised data reduction technique there are network domains which are well separable in low dimensions (Figure \ref{F:fig4}).
Complementing this with the appropriate nonlinear supervised learning algorithm and the right choice of parameters and scores described in the methodology one can nominate a few  “well behaving” networks.
These are the metabolic-, cheminformatics-, animal interaction-, affiliation-, Facebook-, retweet- and tissue networks.
The classification score of these CNDs exceeds  0.7 in dimensions less than three (Figure \ref{F:fig5}B) suggesting the existence of characteristic structural properties.
The associated measures, however, are not unique but replaceable with strongly correlating ones. Also in most of the cases, except for Facebook networks, the best measures do not stand out clearly from the rest, rather the score values are gradually decreasing (Figure \ref{F:fig6}).
Note here also that the definition of relevant features is dimension dependent.
In three dimensions one can get completely different leading features than in two dimensions, leading to the underperformance of the original forward feature selector (Figure \ref{F:fig5}A).

\section{Conclusions}
Many emergent phenomena in complex systems can be understood through the mathematical indicators of the underlying network topology.
Each complex system has its own characteristics and these are connected in nontrivial ways to the structural properties of the associated network.
The preference of authors for certain  mathematical indicators over others for describing complex networks often lacks clear justification and appears to follow changing trends.
One of the aims of the current work is to capture these measures in action, and investigate their true relevance.
We started from the premise that the defining topological measures of a CND should perform well when employed by a classification algorithm.
Multiple previous research on network classification were based on fewer network domains and performed the classification using an arbitrarily chosen subspace of network measures \cite{bib42, bib43}.
In the current work a more heterogeneous dataset and a much more complete feature set is used.
A consistently constructed machine-learning methodology was applied on topological measures extracted from a large number of real networks.
Information on the directionality of edges, weights, temporality, signs, multi-edges and multiple components were ignored.
Bipartite networks were replaced by one-mode projections. The two-dimensional output of unsupervised data reduction algorithms, like the popular linear PCA and the non-linear t-SNE, indicate  strongly nonlinear relationships between the network measures and producing a very skewed and complex distribution of points in the feature space.
Correlation tests suggest that these relationships are also highly domain dependent thus relevant features should only  be defined relative to individual network domains.
A modified forward selection method was applied which combines the advantages of the two alternative methods in machine-learning, forward-selection and backward-elimination.
Our results show that many network groups are structurally distinguishable using their raw topology, and up to three mathematical descriptors.
These network domains include metabolic-, cheminformatics-, animal interaction-, affiliation-, Facebook-, retweet- and tissue networks. Future research should compare the relevant measures obtained here with the measures determined by null models.
Contrary to previous results the existence of a universal feature set which can identify any type of real network cannot be confirmed.
The prominence of leading features is also domain dependent. For instance, Facebook networks have two relevant features, while metabolic networks are equally separable from the rest based on a large number of  features.
Note, however, that this result should be taken with a grain of salt.
Obtaining a high classification score does not exclude the overlap of the domains in the feature space.
If a class is densely crammed into a small subspace it becomes easily “recognizable” even though other classes populate the same region.
This can lead to spurious results for redundant datasets. For instance the genome scale metabolic networks are the full biochemical maps of different organisms and are constructed from metabolic pathways which are common among many phenotypes, resulting in redundancy of this type of networks. Also the classification of CNDs is not always clear in the repositories.
Social networks for example may contain Facebook and retweet networks while the infrastructure domain may include road and transportation networks.
We found that some groups are poorly distinguishable from each other. This may be due to the scrupulous preprocessing and the limited dataset.
However, if classification is the primary goal one can also take into consideration the ignored degrees of freedom such as directionality or edge weights.
Based on the identified set of features relevant for these domains, one can take this investigation further and search for the forces driving their evolution by considering the optimization of the given topological descriptors or comparing them with null models restricting the important features to come up with other relevant properties. 
According to our reported experience the merger of network science and machine learning provides redoubtable means to scientists.
However, it is still in its infancy and as such full of pitfalls.
In our opinion the rapid development of these two fields will soon provide mature solutions to the problems presented here and will revolutionize the way we view and make use of complex systems. 

\section*{Acknowledgements}
This work was supported by the Collegium Talentum Programme of Hungary.
M. J. acknowledges a STAR UBB excellence bursary from the Council for Research of the Babes-Bolyai University.
A.S.L. is supported by a seed award in science from the Wellcome trust (207799/Z/17/Z).
The authors are grateful to Z. Néda, M. Ercsey-Ravasz, B. Sándor, E. Szathmáry, R. Albert and B.Á. Pataki for enlightening discussions and many edifying suggestions.
Furthermore, the work could not be completed without the support of  A. Libál, B. Molnár and L. Varga to access the necessary computational resources. 

\nocite{bib59, bib60, bib61, bib62, bib63, bib64, bib65, bib66, bib67, bib68, bib69, bib70, bib71, bib72, bib73, bib74}

\bibliographystyle{unsrt}  
\bibliography{main}  

\end{document}